%% file: main.tex
\documentclass[acmtog]{acmart}
\usepackage{enumitem}
\copyrightyear{2026}
\acmYear{2026}
\setcopyright{cc}
\setcctype{by-nc-nd}
\acmConference[SA Conference Papers '26]{SIGGRAPH Asia 2026 Conference Papers}{December 01--04, 2026}{Kuala Lumpur, Malaysia}
\acmBooktitle{SIGGRAPH Asia 2026 Conference Papers (SA Conference Papers '26), December 01--04, 2026, Kuala Lumpur, Malaysia}
\acmDOI{10.1145/3829340.3842155}
\acmISBN{979-8-4007-2842-6/2026/12}

\begin{document}

\title[AnimalLift]{AnimalLift: Reconstructing Animatable 3D Animals from a Single Image by Learning Canonical Shape, Texture, and Fur Maps}

\author{Chunyi Sun}
\orcid{0009-0002-7144-1332}
\affiliation{%
  \institution{Australian National University}
  \city{Canberra}
  \state{ACT}
  \country{Australia}}
\email{chunyi.sun@anu.edu.au}

\author{Ruyi Zha}
\affiliation{%
  \institution{Australian National University}
  \city{Canberra}
  \state{ACT}
  \country{Australia}}
\email{ruyi.zha@anu.edu.au}

\author{Weijian Deng}
\affiliation{%
  \institution{Australian National University}
  \city{Canberra}
  \state{ACT}
  \country{Australia}}
\email{weijian.deng@anu.edu.au}

\author{Junlin Han}
\affiliation{%
  \institution{University of Oxford}
  \city{Oxford}
  \country{United Kingdom}}
\email{junlinhcv@gmail.com}

\author{Dylan Campbell}
\affiliation{%
  \institution{Australian National University}
  \city{Canberra}
  \state{ACT}
  \country{Australia}}
\email{dylan.campbell@anu.edu.au}

\author{Stephen Gould}
\affiliation{%
  \institution{Australian National University}
  \city{Canberra}
  \state{ACT}
  \country{Australia}}
\email{stephen.gould@anu.edu.au}

\renewcommand\shortauthors{Sun et al.}

\begin{abstract}
Reconstructing a fully animatable 3D animal from a single image remains challenging because animation-ready assets require not only plausible geometry, but also a unified topology, editable appearance, and fur representations compatible with deformation and simulation. Existing image-to-3D approaches often rely on implicit or loosely structured representations that are difficult to rig or edit, while parametric animal models support animation but cannot capture detailed texture and fur appearance.
We present AnimalLift, a framework for reconstructing structured, animation-compatible 3D animal assets with explicit fur from a single image. Our method lifts an input image into a shared canonical space with a consistent topology and UV parameterization across the dataset, enabling joint prediction of canonical geometry, texture, and fur in a unified feed-forward architecture. A key component of our representation is a UV-aligned fur map that encodes strand geometry in a surface-aligned canonical domain, allowing explicit fur reconstruction compatible with mesh deformation and fur simulation.
To train the model, we introduce a procedural data generation pipeline that provides large-scale supervision with aligned geometry, texture, and fur across diverse animal species and appearances. Experiments on synthetic and real-world datasets demonstrate strong reconstruction quality and generalization across animal categories. Beyond reconstruction, our structured representation directly supports downstream applications including animation, pose transfer, fur editing, and simulation-compatible rendering.

\end{abstract}



\begin{teaserfigure}
    \centering
    \includegraphics[width=1\textwidth]{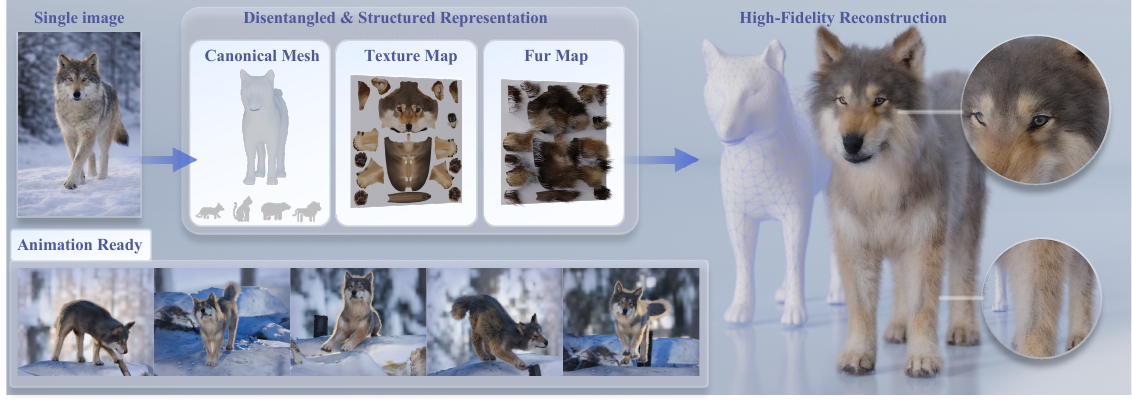}
    \caption{
    From a single animal image, our method reconstructs a canonical animatable asset for a shared topology with UV-aligned texture and an explicit fur representation.
    The disentangled canonical representation enables high-quality appearance reconstruction together with downstream applications including animation, grooming, editing, and physics-based fur simulation.
    }
    \Description{A wolf photograph is mapped to a canonical mesh, a UV texture map, and a fur map. Renderings show the reconstructed wolf, fur close-ups, and a sequence of animated poses.}
    \label{fig:main_vis}
\end{teaserfigure}

\maketitle

\input{sec/1_intro}    
\input{sec/2_related_work}
\input{sec/3_dataset}

\input{sec/4_method}

\input{sec/5_results}
\input{sec/6_conclusion}

\clearpage
\begin{figure*}[t]
    \centering

    \includegraphics[width=\textwidth]{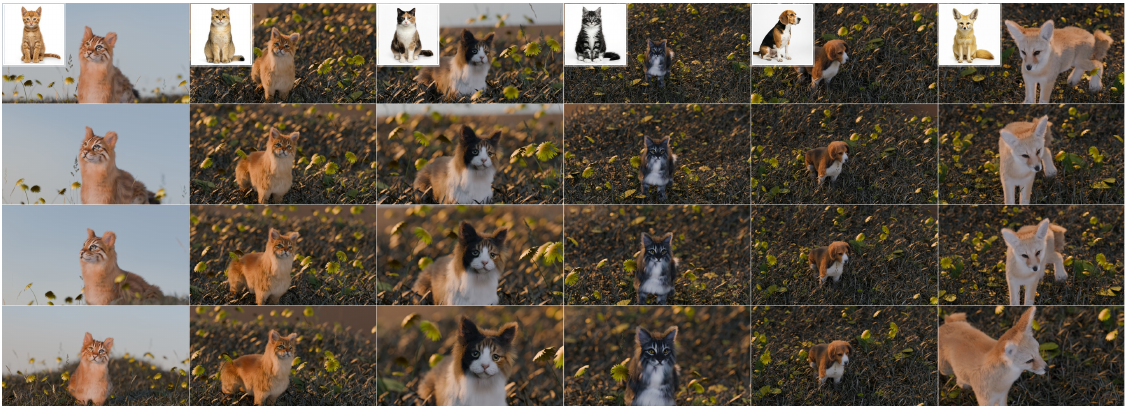}
    \caption{\textbf{3D animals reconstructed from a single input image.}
    Results are rendered in natural scenes to reveal appearance under diverse lighting,
    with zoom-ins highlighting fur detail and structure. Our method generalizes across
    different species while preserving geometry, identity, and coherent hair patterns.}
    \Description{Four rows show animal photographs and their reconstructed assets placed in outdoor scenes. Insets enlarge fur details and local appearance.}
    \label{fig:demo2}

    \begin{minipage}[t]{0.48\textwidth}
        \centering
        \includegraphics[width=\linewidth]{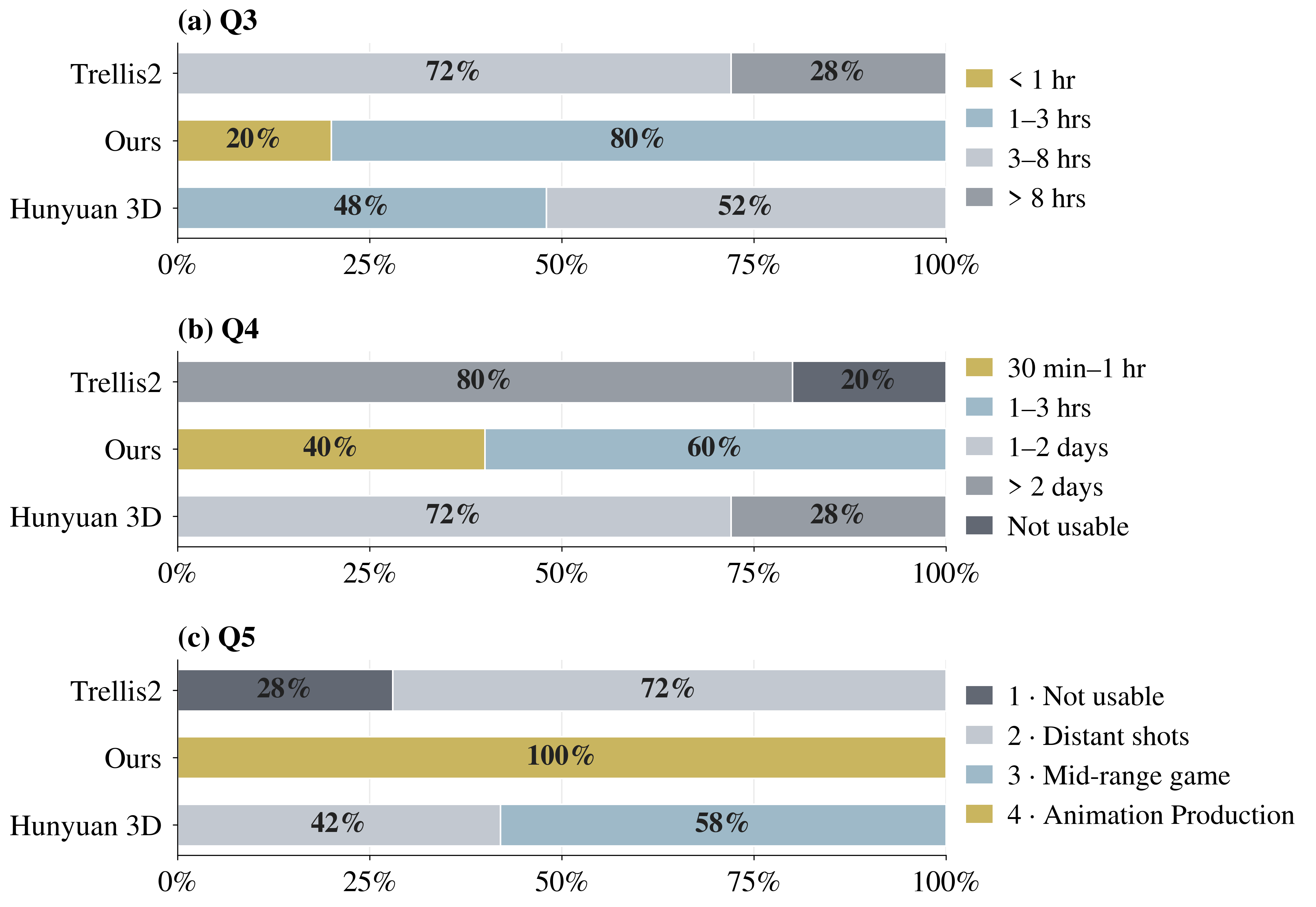}
        \caption{\textbf{Expert ratings of cleanup effort and production usability.}
        Rating distributions for (a) static-render cleanup effort, (b) manual effort
        for 10-second walking-animation preparation, and (c) the highest quality level
        reachable after one hour of cleanup.}
        \Description{Three groups of horizontal stacked bars compare Trellis 2, AnimalLift, and Hunyuan-3D on static-render cleanup time, walking-animation preparation time, and quality after one hour of cleanup.}
    \label{fig:user_study}
    \end{minipage}
    \hfill
        \begin{minipage}[t]{0.48\textwidth}
        \centering
        \includegraphics[width=0.7\linewidth]{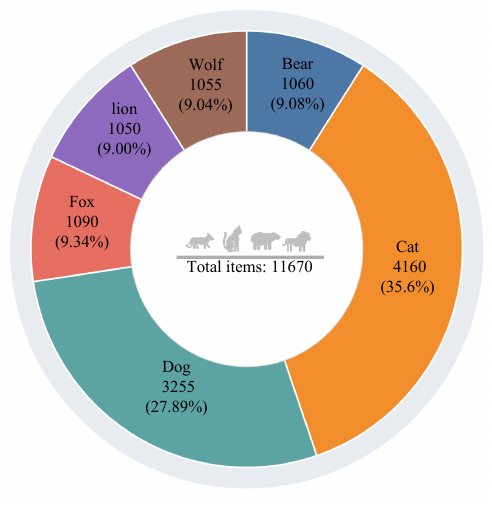}
        \caption{\textbf{Dataset statistics.}
        Distribution of animal categories in our dataset, showing the number of examples
        for each animal class used in training and evaluation.}
        \Description{A donut chart shows the distribution of bear, cat, dog, fox, lion, and wolf examples. Cats and dogs form the largest portions.}
    \label{fig:dataset_stat}
    \end{minipage}

    \begin{minipage}[t]{0.48\textwidth}
        \centering
        \includegraphics[width=0.8\linewidth]{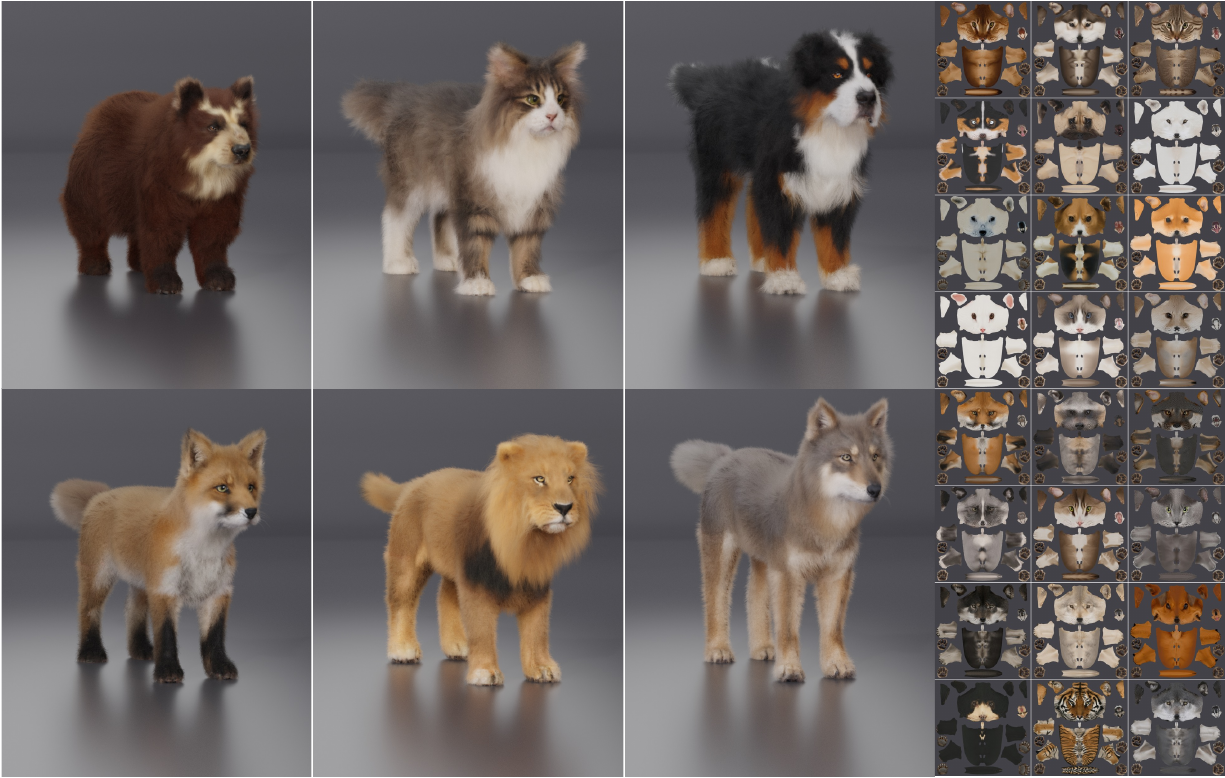}
        \caption{\textbf{Dataset visualization.}
        Left: representative 3D animal models from each class in our dataset. Right: diverse UV texture maps
        that share a common layout across the animals.}
        \Description{Six representative furry animal models appear beside a grid of UV texture maps with a shared anatomical layout.}
    \label{fig:dataset}
    \end{minipage}
    \hfill
    \begin{minipage}[t]{0.48\textwidth}
        \centering
        \includegraphics[width=\linewidth]{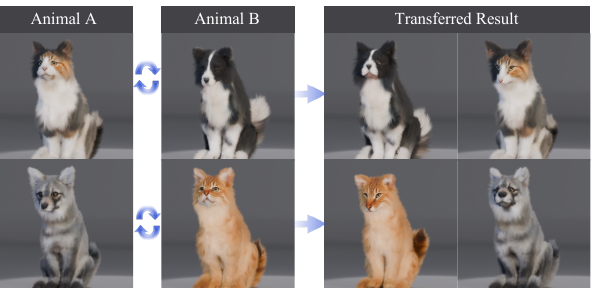}
        \caption{\textbf{Fur and texture transfer across animals.}
        Given two different animal models, our method swaps their fur geometry and texture
        appearance while preserving the underlying body shape and identity of each target.}
        \Description{Two pairs of animals are shown before and after swapping fur and texture. The transferred results retain the target body shapes while exchanging coat appearance.}
    \label{fig:fur_transfer}
    \end{minipage}

\end{figure*}

\clearpage
\bibliographystyle{ACM-Reference-Format}
\bibliography{sample-bibliography}

\end{document}

%% file: sec/1_intro.tex
\section{Introduction}

Transforming a single image of an animal into a fully animatable 3D asset with realistic fur remains a long-standing challenge in computer graphics. Beyond recovering plausible surface geometry, animation-ready assets must support rigging, deformation, editable appearance, and realistic fur motion while maintaining consistent correspondences under articulated motion. Producing such assets traditionally requires extensive manual modeling, texturing, grooming, and rigging, making high-quality digital animal creation both time-consuming and expensive.

Recent image-to-3D reconstruction methods~\cite{hunyuan3d2025hunyuan3d,tochilkin2024triposr,trellis,hong2024lrm} can generate visually compelling 3D assets from single images. However, they often rely on implicit fields, point-based representations, or unconstrained meshes. These outputs lack shared topology, canonical correspondences, and editable surface parameterizations, making them difficult to rig, animate, or directly reuse in standard graphics pipelines. For articulated animals, shared topology is also important for learning. It assigns consistent meanings to vertices, body regions, and UV coordinates across instances, allowing geometric and appearance supervision to be aggregated in a common domain.

Parametric animal models~\cite{Zuffi:CVPR:2017,rueegg2022barc,ruegg2023bite,li2021hsmal,zuffi2024varen} provide this structure and support animation, but their low-dimensional formulations struggle to capture fine-scale geometry, detailed textures, and high-frequency fur appearance. This highlights a key gap: animatable animal assets require a structured representation that jointly models geometry, texture, and fur under canonical correspondences. Such a representation should preserve dense correspondences, disentangle appearance from pose, and remain compatible with deformation and simulation. In particular, fur should be modeled as an explicit surface-aligned structure that deforms consistently with the underlying animal geometry, rather than as a view-dependent rendering effect.

Our key insight is that both surface appearance and fur can be represented in a shared UV-aligned canonical space with consistent topology and parameterization. In particular, we model fur as a UV-aligned fur map that encodes strand geometry in a surface-aligned canonical domain, enabling explicit fur reconstruction compatible with mesh deformation and fur simulation.
Motivated by this, we present AnimalLift, a framework for reconstructing structured, animation-compatible 3D animal assets with explicit fur from a single image. Our method lifts an input image into a shared canonical space with consistent topology and UV parameterization, enabling joint prediction of canonical mesh geometry, texture, and fur in a unified feed-forward architecture. By reconstructing a canonical asset rather than a posed surface, our approach produces outputs with dense correspondence and disentangled appearance that can be directly animated using standard deformation models.

To enable learning of such structured outputs, we introduce a procedural data generation pipeline that provides aligned supervision of geometry, texture, and fur in canonical space. The pipeline synthesizes diverse animal species, shapes, textures, and fur appearances while maintaining shared topology and UV correspondence.

Experiments on both synthetic and real-world datasets demonstrate strong reconstruction quality and generalization across animal categories. More importantly, our representation directly supports downstream graphics applications including animation, pose transfer, fur editing, and simulation-compatible fur rendering.

In summary, our contributions are:
\begin{enumerate}
    \item a canonical UV-aligned representation for jointly modeling animal geometry, texture, and explicit fur in a structured animation-compatible space;
    
    \item a unified feed-forward framework for reconstructing animatable 3D animal assets with explicit fur; 
    
    \item a procedural data generation pipeline that provides aligned supervision of geometry, texture, and fur across diverse animal species and appearances.
    
\end{enumerate}

\noindent Code, trained models, and dataset are available at\newline \url{https://huggingface.co/datasets/Chunyi99/AnimalLift}.

%% file: sec/2_related_work.tex
\section{Related Work}
\label{sec:formatting}

\paragraph{Image-to-3D Reconstruction} A large body of work studies how to recover 3D shape from images, especially with neural rendering and generative models. Neural radiance fields~\cite{mildenhall2021nerf} learn implicit radiance fields from multiple views, and subsequent NeRF-based methods such as DreamFusion~\cite{poole2022dreamfusion}, Magic3D~\cite{lin2023magic3d}, and Zero-1-to-3~\cite{liu2023zero} leverage pretrained generative models to synthesize 3D content from a single image or text prompt. Very recent systems~\cite{yang2024hunyuan3d,tochilkin2024triposr,trellis,li2024instant3d,wei2024meshlrm,xu2024grm} further improve efficiency and visual quality with feed-forward generation. Despite these advances, most of these methods rely on implicit or loosely structured representations, without consistent topology or dense correspondences. So, their outputs can be difficult to rig, deform, or reuse in downstream graphics pipelines.

\paragraph{Parametric Animal Models} Parametric models provide a shared topology that decomposes shape, pose, and appearance, making them well suited for deformation and animation. SMAL model~\cite{Zuffi:CVPR:2017} introduces a statistical quadruped shape space and underpins most subsequent animal models, analogous to SMPL~\cite{loper2023smpl} for humans. 
Follow-up works extend to specific species and settings, including BARC~\cite{rueegg2022barc} and BITE~\cite{ruegg2023bite} for dogs, hSMAL~\cite{li2021hsmal} and VAREN~\cite{zuffi2024varen} for horses, and SMAL+~\cite{zuffi2024awol}, an extended SMAL learned from additional scans. These models are often used as structural priors in reconstruction pipelines, but their low-dimensional formulations capture coarse shape while struggling with fine geometric detail, high-frequency texture, and fur.

\paragraph{Animal Reconstruction} To better handle articulated objects, several works focus on category-specific reconstruction with stronger priors. Methods based on SMAL~\cite{Zuffi:CVPR:2017} and its applications~\cite{zuffi2018lions,zuffi2019three} estimate animal shape and pose from images. More recent approaches learn deformation and motion directly from videos~\cite{yang2022banmo,sun2024ponymation,luo2022artemis,sabathier2024animalavatars,wu2023dove,sinha2023commonpets,lei2024gart} and images~\cite{wu2023magicpony,li2024learning}. While these methods capture articulation and coarse geometry, they usually operate in posed space and do not explicitly disentangle canonical shape, texture, and appearance. As a result, the reconstructed outputs are often tied to a specific instance and are not directly reusable.

\paragraph{Automatic Rigging} Several works predict skeletons and skinning weights to rig generated 3D assets~\cite{xu2020rignet,liu2025riganything,zhang2025one}, enabling downstream deformation and animation. However, they are typically applied after geometry reconstruction and can be sensitive to mesh consistency and quality. In contrast, our shared-topology assets are rig-compatible by construction.

\paragraph{Hair and Fur Modeling} Classical approaches rely on strand-based grooming systems (\textit{e.g.}, production pipelines in Blender or Houdini), while learning-based methods~\cite{zhou2018hairnet,zheng2023hairstep,rosu2022neural,he2025perm,zhou2023groomgen,luo2024gaussianhair} aim to predict hair geometry from images. More recent work targets animal fur specifically: NeuralFur~\cite{sklyarova2026neuralfur} reconstructs strand-based fur from multi-view images. These methods usually rely on multiview images and do not provide a representation aligned with a canonical surface that can be directly used for animation. Related surface-appearance and neural-texture representations~\cite{chang2026lito,oechsle2019texturefields,thies2019deferred} jointly encode geometry and appearance but target view-dependent effects rather than explicit, editable fur. In addition, unlike human hair, animal fur often defines the outer silhouette and can be strongly decoupled from the underlying body shape between species. This makes fur generation challenging, requiring both surface consistency and accurate recovery of the visible fur volume.

%% file: sec/3_dataset.tex
\begin{figure*}[!t]
    \centering
    \includegraphics[width=0.9\textwidth]{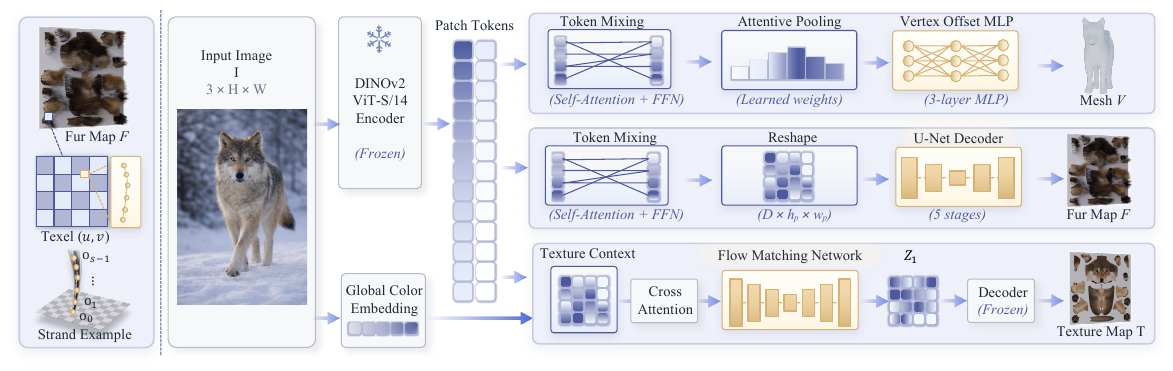}
    \caption{\textbf{Overview of the proposed method.} Left: each occupied UV texel represents a surface-rooted fur strand and stores root-relative 3D offsets for its strand points. Right: the input image is encoded by a partially frozen DINOv2 ViT-S/14 backbone and a global color encoder, with separate branches predicting the canonical mesh, UV texture map, and UV-aligned fur map.}
    \Description{The diagram aligns strand offsets with UV texels. A DINOv2 image encoder and global color encoder condition separate vertex-offset, fur-map, and texture-flow branches, producing canonical geometry, fur, and texture.}
    \label{fig:method}

\end{figure*}

\section{Canonical Data Generation}
\label{sec:dataset}

AnimalLift requires supervision that is aligned across geometry, texture, and fur, which is not available from real images. We therefore synthesize animal assets in a shared canonical space, where all instances have consistent topology, skeleton, and UV parameterization, but vary in shape, appearance, fur, pose, view, and lighting. This produces paired training data from posed single-image observations to canonical animatable assets. 

\paragraph{Canonical Shape Space.}
We start from a shared-topology quadruped template derived from SMAL~\cite{Zuffi:CVPR:2017}, preserving mesh connectivity and skeletal correspondence across all assets. We also manually design a canonical UV layout that is spatially uniform, semantically organized, and visually interpretable as an unfolded animal body. Corresponding anatomical regions, such as the head, torso, limbs, ears, and tail, occupy consistent UV regions across all instances. This UV design is critical for both dense correspondence and texture editing.

To cover anatomical diversity, we group animals by coarse morphology, including dogs, cats, foxes, wolves, bears, and lions. For each group, artists sculpt representative breed- or species-level meshes with shared topology, skeleton correspondence, and UV layout. All meshes are expressed in the canonical shape pose, following the SMAL convention where each articulated joint has zero relative rotation with respect to its parent joint. We then augment these bases with topology-preserving interpolation, shape mixing, and semantic shape keys for attributes such as muzzle length, ear shape, limb length, chest width, tail shape, and body proportions. Thus, all generated shapes retain dense vertex correspondence and share a common rig and UV domain.

\paragraph{Canonical Texture and Fur Generation.}
Textures are generated in the canonical UV domain. For each canonical shape, we first create a complete base UV texture using online texturing systems, including Vast-Tripo~\cite{tripo2026online} and Hunyuan3D Studio~\cite{hunyuan3dstudio}, conditioned on the geometry and species- or morphology-matched references. We then augment these templates using Qwen-Image-Edit~\cite{wu2025qwen} directly in UV space, varying coat color, markings, stripes, spots, facial patterns, and local fur appearance. Since the UV atlas is semantically organized, these edits remain localized to meaningful anatomical regions while preserving global body consistency. These external texturing tools~\cite{tripo2026online,hunyuan3dstudio,wu2025qwen} are used only for offline data generation and are not part of the trained model or inference pipeline.

Fur is generated from shared canonical grooming templates defined in the same UV domain. Artists manually create the base groom once, and derive representative templates by annotating density, length, and curliness maps over semantic body regions. These maps include smooth transitions in sparse-fur areas such as the feet, inner limbs, and underarms. For each animal, we adapt the shared maps according to species and morphology, then apply stochastic UV-space perturbations to produce instance-specific variation. Creating the canonical templates (topology, UV, rig, skinning, shape keys, and base groom) is a one-time initialization per morphology; all subsequent training samples are then generated procedurally without further manual intervention. Strand-based fur is procedurally generated from these maps and rooted on the canonical mesh. Each strand root is associated with a UV coordinate, allowing the groom to be stored and supervised in canonical UV space. We parameterize each strand by resampling it into a fixed number of points and storing the root-relative 3D offsets of the remaining points in the canonical coordinate system.

\paragraph{Rendering and Pose Augmentation.}
To generate training observations, each canonical asset is articulated using poses sampled from a category-level pose pool derived from SMAL-compatible pose estimation~\cite{ruegg2023bite} on posed animal images. The asset is deformed with linear blend skinning and rendered under randomized viewpoints, camera distances and HDR environment maps. Pose, camera, and illumination are used only to synthesize the input image; all targets remain canonical.

\paragraph{Canonical Supervision and Representation.}
Each training sample consists of a rendered image and its canonical asset: $(\mathbf{I}, \mathbf{V^*}, \mathbf{T^*}, \mathbf{F^*}, \mathbf{D^*})$, where $\mathbf{I} \in \mathbb{R}^{3 \times H \times W}$ is the input image,
$\mathbf{V^*} \in \mathbb{R}^{N \times 3}$ is the canonical mesh,
$\mathbf{T^*} \in \mathbb{R}^{3 \times H_t \times W_t}$ is the UV texture,
$\mathbf{D^*} \in \mathbb{R}^{H_f \times W_f}$ is the fur density map, and
$\mathbf{F^*} \in \mathbb{R}^{H_f \times W_f \times (S-1) \times 3}$ is the UV-aligned fur tensor. The fur tensor stores root-relative strand geometry. For each occupied UV cell, the strand root lies on the corresponding canonical surface point, and $\mathbf{F^*}$ stores offsets from the root to the remaining resampled strand points; non-fur cells are indicated by $\mathbf{D^*}$. 

In total, our dataset covers six animal categories, including dogs, cats, foxes, wolves, bears, and lions, and contains 11K paired samples of posed single-image observations and canonical animatable assets. The dataset is split at the breed level so that no breed appears in both training and evaluation. A per-stage breakdown of which components are reusable, automatic, and manual is provided in the supplementary material, together with additional generation details and statistics (Fig.~\ref{fig:dataset_stat}, supplementary material Section 1).

%% file: sec/4_method.tex
\section{AnimalLift}
\paragraph{Method Overview.} Given a single RGB image $\mathbf{I} \in \mathbb{R}^{3 \times H \times W}$ of a quadruped animal, our goal is to reconstruct a canonical animatable asset comprising a mesh $\mathbf{V}^* \in \mathbb{R}^{N \times 3}$, a UV texture map $\mathbf{T}^* \in \mathbb{R}^{3 \times H_t \times W_t}$, and a UV fur map $\mathbf{F}^* \in \mathbb{R}^{C_f \times H_f \times W_f}$. The architecture consists of a shared DINOv2 encoder~\cite{dinov2} followed by three specialized heads: a vertex offset head $f_{\text{vert}}$, a fur map head $f_{\text{fur}}$, and a texture flow-matching network $f_{\text{tex}}$. All outputs are defined in the canonical space, enabling direct dense supervision from the dataset described in Section~\ref{sec:dataset}. The texture map $\mathbf{T}^*$ stores RGB values, whereas the fur map $\mathbf{F}^*$ stores UV-aligned strand attributes (root-relative 3D offsets and a per-cell density).

\subsection{Shared Image Encoder}

We adopt a DINOv2 ViT-S/14~\cite{dinov2} encoder $\mathcal{E}_{\text{dino}}$ as the shared backbone. The encoder is mostly frozen during training, except for the last two transformer blocks and final layer norm, which are fine-tuned at $0.1\times$ the base learning rate. It maps a normalized input image to final patch tokens $\mathbf{H}=\mathcal{E}_{\text{dino}}(\mathbf{I}) \in \mathbb{R}^{L \times D}$, where $L=(H/14)\times(W/14)$ and $D=384$.

The geometry and fur heads operate directly on $\mathbf{H}$. For texture synthesis, we build a richer conditioning context that also captures absolute color and multi-scale appearance cues, which are partially suppressed in the final DINOv2 tokens. We concatenate a global color token, intermediate DINOv2 tokens, and final tokens: $
\mathbf{C}=[\,\mathbf{c}_{\text{color}},\mathbf{H}_{\text{inter}},\mathbf{H}\,]\in\mathbb{R}^{L'\times D},$
where $\mathbf{c}_{\text{color}}\in\mathbb{R}^{1\times D}$ and $\mathbf{H}_{\text{inter}}\in\mathbb{R}^{L_{\text{inter}}\times D}$.

The color token is produced by $\mathcal{E}_{\text{color}}$. It encodes per-channel color statistics, including histograms with $K=16$ differentiable triangular-kernel bins, mean, standard deviation, and quantiles at $\{0.05,0.25,0.5,0.75,0.95\}$. These statistics are computed in RGB, matching the space of the predicted texture. These statistics are projected through a two-layer MLP to obtain $\mathbf{c}_{\text{color}}$. The MLP is trained jointly with the flow-matching network and provides explicit coat-color information. We obtain $\mathbf{H}_{\text{inter}}$ from ViT layers 3 and 7. Each layer uses a linear projection and LayerNorm head, both trained with the flow network. These intermediate tokens provide mid-level texture and edge cues that complement the final DINOv2 tokens. The tokens are spatially pooled to a fixed $16\times16$ grid, yielding $L_{\text{inter}}=2\times16^2=512$ tokens.

\subsection{Geometry Reconstruction}

The vertex offset head $f_{\text{vert}}$ applies $K_{\text{vert}} = 2$ token mixing blocks to $\mathbf{H}$, each consisting of multi-head self-attention followed by a feed-forward layer with residual connections, producing $\tilde{\mathbf{H}}_{\text{vert}} = \mathrm{TokenMix}^{K_{\text{vert}}}(\mathbf{H}) \in \mathbb{R}^{L \times D}$. The mixed tokens are aggregated via learned attentive pooling, $\mathbf{g} = \sum_{l=1}^{L} \mathrm{softmax}(s(\tilde{\mathbf{H}}_{\text{vert}}))_l \tilde{\mathbf{H}}_{\text{vert},l} \in \mathbb{R}^{D}$, where $s : \mathbb{R}^{D} \rightarrow \mathbb{R}$ is a two-layer scoring MLP applied per token. This pooling is used only for geometry; the fur head instead keeps a dense per-cell UV field (below) to preserve spatial structure. The resulting global descriptor is passed through a three-layer MLP $\Delta\mathbf{V} = f_{\text{vert}}(\mathbf{g}) \in \mathbb{R}^{N \times 3}$ with GELU activations, dropout, and zero-initialized output layer. The canonical mesh is recovered as $\mathbf{V} = \bar{\mathbf{V}} + \Delta\mathbf{V}$, where $\bar{\mathbf{V}} \in \mathbb{R}^{N \times 3}$ is the mean template computed by averaging vertex positions across all training instances. The geometry head is supervised with a composite loss:
\begin{equation}
    \mathcal{L}_{\text{mesh}} = \mathcal{L}_{\text{vert}}
    + \lambda_e \mathcal{L}_{\text{edge}}
    + \lambda_l \mathcal{L}_{\text{lap}}
    + \lambda_r \mathcal{L}_{\text{reg}}
    \label{eq:mesh_loss}
\end{equation}
where $\mathcal{L}_{\text{vert}} = \frac{1}{N}\|\mathbf{V} - \mathbf{V}^*\|_1$ measures per-vertex accuracy, $\mathcal{L}_{\text{edge}}$ penalizes differences in predicted versus ground-truth edges, $\mathcal{L}_{\text{lap}}$ encourages smooth deformations by penalizing deviation of the predicted mesh Laplacian from that of the template, and $\mathcal{L}_{\text{reg}} = \frac{1}{N}\|\Delta\mathbf{V}\|_1$ penalizes large offsets. 

\subsection{Fur Reconstruction}

The fur head $f_{\text{fur}}$ applies $K_{\text{fur}} = 2$ token mixing blocks to $\mathbf{H}$ and reshapes the output into a 2D spatial feature map $\mathbf{P} \in \mathbb{R}^{D \times h_p \times w_p}$, where $h_p = w_p = \sqrt{L}$. This reshaping is valid because DINOv2 patch tokens maintain their spatial ordering, so the 2D layout of $\mathbf{P}$ corresponds directly to the spatial layout of the input image. A lightweight UNet-style decoder with five progressive upsampling stages and channel dimensions $(256, 192, 128, 96, 64)$ produces the dense UV fur map jointly predicting strand offsets and strand density: $\hat{\mathbf{F}} = f_{\text{fur}}(\mathbf{P}) \in \mathbb{R}^{(C_f + 1) \times H_f \times W_f}$, where the first $C_f = (S-1) \times 3 = 189$ channels encode root-relative 3D offsets of $S-1 = 63$ resampled strand points per UV cell, and the final channel is a density map $\hat{\mathbf{D}} \in \mathbb{R}^{H_f \times W_f}$ recording the number of child strands to interpolate at each UV cell. Both outputs share the same decoder and are predicted jointly. Training supervises the offset channels with a foreground-weighted masked $\ell_1$ loss $\mathcal{L}_{\text{fur}}$ restricted to valid UV cells under the fur validity mask $m(\mathbf{u}) \in \{0,1\}$, upweighting occupied cells by $w_{\text{fg}} = 2$ relative to background, and supervises $\hat{\mathbf{D}}$ with an $\ell_1$ loss against the ground-truth strand count map $\mathbf{D}^*$.

\subsection{Texture Synthesis via Flow Matching}

Instead of directly regressing UV textures, we employ a flow-match\-ing \cite{flowmatching,flowmatching2} formulation that better captures the multimodal distribution of high-frequency animal appearance, while remaining substantially more lightweight and controllable than finetuning large-scale image diffusion models designed for unconstrained image synthesis. We apply flow matching only to texture, whose completion is inherently multimodal (occluded regions and fine markings admit many plausible RGB completions), so deterministic regression tends to blur it (Table~\ref{tab:ablation}). Geometry and fur instead require topology/rig consistency and root-attached strand coherence, which deterministic regression enforces naturally; adding flow matching to fur gave no perceptual gain at higher cost.

We first train a convolutional autoencoder $(\mathcal{E}_{\text{tex}}, \mathcal{D}_{\text{tex}})$ that maps $\mathbf{z} = \mathcal{E}_{\text{tex}}(\mathbf{T}) \in \mathbb{R}^{8 \times 64 \times 64}$ and $\hat{\mathbf{T}} = \mathcal{D}_{\text{tex}}(\mathbf{z}) \in \mathbb{R}^{3 \times 1024 \times 1024}$, achieving $16\times$ spatial compression. The autoencoder is trained independently with a masked reconstruction loss upweighting the UV-occupied foreground region and a latent regularization term $\lambda_{\text{reg}}\|\mathbf{z}\|_2^2$. Concretely, $\mathcal{E}_{\text{tex}}$ and $\mathcal{D}_{\text{tex}}$ form a mirrored convolutional encoder--de\-coder with residual blocks and strided down-/up-sampling between $1024^2$ RGB texture and  $8\times64^2$ latent, and $f_{\text{tex}}$ is a UNet with symmetric down/up stages and residual blocks operating on this latent.

Given a ground-truth latent $\mathbf{z}_0 = \mathcal{E}_{\text{tex}}(\mathbf{T}^*)$, we define a linear interpolation path $\mathbf{z}_t = (1-t)\boldsymbol{\epsilon} + t\mathbf{z}_0$ with $\boldsymbol{\epsilon} \sim \mathcal{N}(0,\mathbf{I})$ and $t \sim \mathcal{U}(0,1)$, yielding target velocity $\mathbf{v}^* = \mathbf{z}_0 - \boldsymbol{\epsilon}$. The flow network predicts this velocity conditioned on the noisy latent, the timestep, and the augmented texture context $\mathbf{C}$ as $\hat{\mathbf{v}} = f_{\text{tex}}(\mathbf{z}_t,\, t,\, \mathbf{C}) \in \mathbb{R}^{8 \times 64 \times 64}$, where $\mathbf{C}$ is injected via cross-attention at the two deepest UNet levels. To enable classifier-free guidance~\cite{ho2022classifier}, the condition $\mathbf{C}$ is replaced by a zero vector with probability $p_{\text{drop}} = 0.10$ during training. At inference, texture is synthesized by Euler integration over $N_{\text{step}} = 50$ steps with guidance scale $s = 3.0$, using $\hat{\mathbf{v}}_{\text{cfg}} = \hat{\mathbf{v}}_\varnothing + s\,(\hat{\mathbf{v}}_{\mathbf{C}} - \hat{\mathbf{v}}_\varnothing)$, followed by decoding $\hat{\mathbf{T}} = \mathcal{D}_{\text{tex}}(\hat{\mathbf{z}}_1) \in \mathbb{R}^{3 \times 1024 \times 1024}$. The flow loss $\mathcal{L}_{\text{tex}}$ is a foreground-weighted MSE between $\hat{\mathbf{v}}$ and $\mathbf{v}^*$ under the downsampled UV occupancy mask.

\subsection{Training}

Training proceeds in two stages. First, the texture autoencoder $(\mathcal{E}_{\text{tex}}, \mathcal{D}_{\text{tex}})$ is trained independently. Second, the autoencoder is frozen and $f_{\text{vert}}$, $f_{\text{fur}}$, and $f_{\text{tex}}$ are trained jointly. Joint training lets the three heads share a single image-conditioned backbone and one forward pass over common cues (silhouette, species, proportion, and fur direction), which is more efficient than training three separate networks while incurring no loss in quality. The geometry and fur heads receive clean final DINOv2 tokens $\mathbf{H}$ without condition dropout, while the texture flow network receives the CFG-dropped augmented context $\mathbf{C}$. This separation ensures that geometric prediction is never trained on zeroed conditioning, preserving the quality of shape and fur supervision. The total joint objective is:
\begin{equation}
    \mathcal{L} = \lambda_{\text{geo}}\,\mathcal{L}_{\text{mesh}}
    + \lambda_{\text{fur}}\,\mathcal{L}_{\text{fur}}
    + \lambda_{\text{tex}}\,\mathcal{L}_{\text{tex}}
    \label{eq:total_loss}
\end{equation}
with $\lambda_{\text{geo}} = 10$, $\lambda_{\text{fur}} = 1$, $\lambda_{\text{tex}} = 1$. 

\subsection{Animation and DCC Integration}

A key advantage of our canonical representation is direct compatibility with standard DCC\footnote{DCC refers to digital content creation software, such as Blender, Maya, or Houdini, where artists edit, rig, animate, groom, and simulate 3D assets in production workflows.} pipelines. Since reconstructed animals share topology, skeleton correspondence, and UVs, predicted meshes can be transferred to a common rig for standard skeletal animation. For fur, we use the UV map to locate the root position of each strand on the canonical mesh. The fur geometry is then reconstructed by adding the predicted offsets to the corresponding root positions, producing strand vertices consistently anchored to the canonical surface through their UV-defined roots. The predicted density map $\hat{\mathbf{D}}$ specifies the number of child strands to interpolate per UV cell, which directly maps to Blender's native hair interpolation modifier. The reconstructed guide strands are exported as Blender curve-based hair grooms with UV attributes, mesh bindings, and density information, enabling Blender to automatically interpolate the full groom from the predicted guide strands without any manual re-grooming. This allows native grooming, physics simulation, and rendering within standard production workflows. Blender auto-import details are provided in the supplementary material Section 2.

%% file: sec/5_results.tex
\section{Experiments}

\begin{figure*}[t]
    \centering

    \includegraphics[width=0.90\textwidth]{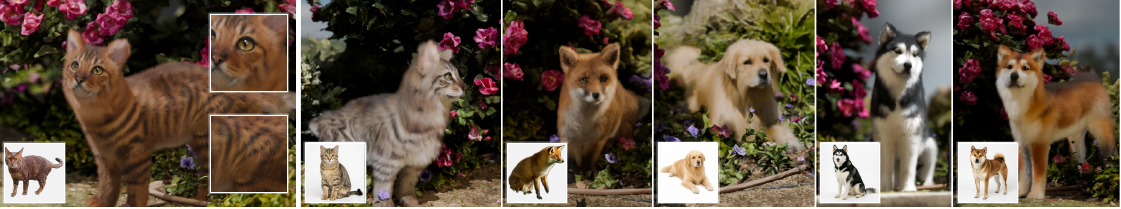}

    \caption{\textbf{3D animals reconstructed from a single input image.} Results are rendered in natural scenes to reveal appearance under diverse lighting, with zoom-ins highlighting fur detail and structure. Our method generalizes across different species while preserving geometry, identity, and coherent hair patterns.}
    \Description{Input photographs accompany reconstructed furry animals rendered in an outdoor scene. Magnified insets show individual strands and local surface appearance.}
    \label{fig:demo1}

\end{figure*}

\begin{figure}[!htbp]
  \includegraphics[width=1\linewidth]{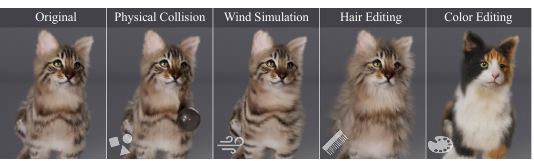}
  \caption{\textbf{Applications.} Our method reconstructs explicit mesh and strand-based hair from a single image, enabling diverse applications. We show physical collision, wind simulation, and controllable hair and color editing. }
  \Description{Five renderings show the original animal and the results of physical collision, wind simulation, hair editing, and color editing.}
    \label{fig:application}
\end{figure}

\begin{figure*}[!t]
    \centering
    \includegraphics[width=0.8\textwidth]{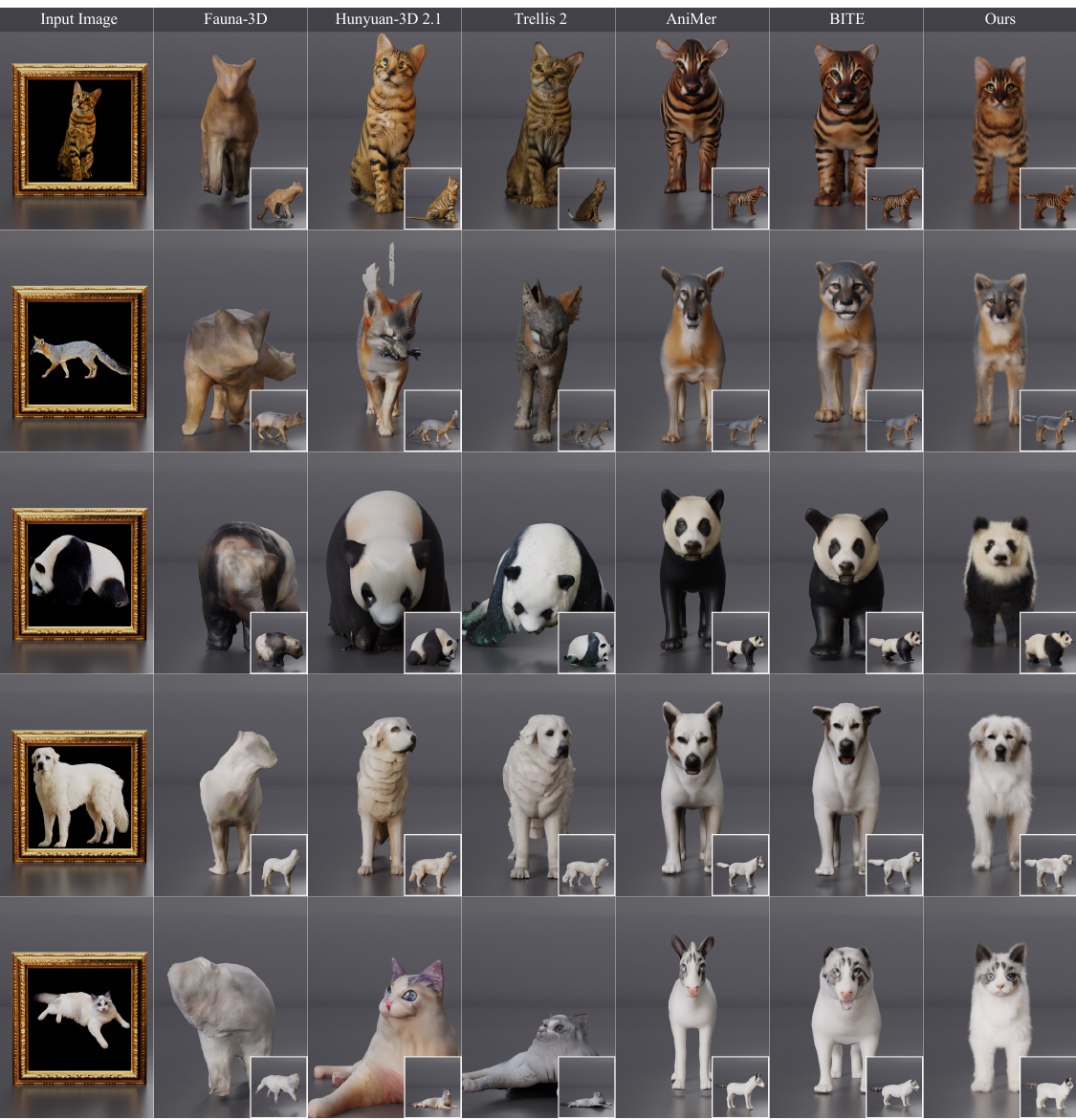}
    \caption{\textbf{Comparison of reconstruction results.} The framed images in the first column denote the guiding reference images. Insets show additional viewpoints. }
    \Description{Five rows compare input animal photographs with reconstructions from Fauna-3D, Hunyuan-3D 2.1, Trellis 2, AniMer, BITE, and AnimalLift. The examples include cats, dogs, and a panda, with additional views in insets.}
    \label{fig:comparsion2}
\end{figure*}

\begin{table*}[t]
\centering
\caption{Comparison of reconstruction quality and output capabilities on synthetic and real datasets. Anim., UV, and Fur indicate support for animation-compatible geometry, UV-aligned textures, and simulation-ready strand-based fur, respectively. Best results are bold, and second-best are underlined.}
\label{tab:main_cmp}
\resizebox{\textwidth}{!}{
\begin{tabular}{lccc|ccccccc|ccccc}
\toprule
& \multicolumn{3}{c|}{\textbf{Capability}} 
& \multicolumn{7}{c|}{\textbf{Synthetic (Test Split)}} 
& \multicolumn{5}{c}{\textbf{Real Images}} \\
\cmidrule(lr){2-4}\cmidrule(lr){5-11}\cmidrule(lr){12-16}
Method
& Anim. & UV & Fur
& CD-L1$\downarrow$
& L1-Hair$\downarrow$
& NIQE$\downarrow$
& LPIPS$\downarrow$
& MV-CLIP$\uparrow$
& FID$_\text{CLIP}\downarrow$
& Patch-FID$_\text{Inc}\downarrow$
& NIQE$\downarrow$
& LPIPS$\downarrow$
& MV-CLIP$\uparrow$
& FID$_\text{CLIP}\downarrow$
& Patch-FID$_\text{Inc}\downarrow$ \\
\midrule
Fauna-3D~\cite{li2024learning}
& -- & -- & --
& 0.111 & -- & 12.51 & 0.869 & 0.749 & 73.06 & 322.13
& 10.91 & 0.639 & 0.757 & 59.62 & 273.39 \\
Hunyuan-3D~2.1~\cite{hunyuan3d2025hunyuan3d}
& -- & \checkmark & --
& \underline{0.103} & -- & 10.04 & 0.785 & 0.804 & 55.61 & 298.84
& 8.67 & 0.641 & \underline{0.882} & 42.09 & 200.11 \\
Trellis 2~\cite{trellis}
& -- & \checkmark & --
& 0.109 & -- & 9.73 & 0.801 & 0.766 & 60.44 & 327.68
& 8.47 & 0.604 & 0.848 & 49.29 & 225.60 \\
AniMer~\cite{lyu2025animer}
& \checkmark & -- & --
& 0.113 & -- & 11.42 & 0.803 & 0.734 & 61.80 & 281.79
& 10.09 & 0.630 & 0.777 & 53.22 & 220.80 \\
BITE~\cite{ruegg2023bite}
& \checkmark & -- & --
& 0.109 & -- & 11.55 & 0.794 & 0.764 & 64.07 & 285.98
& 10.17 & 0.629 & 0.799 & 48.80 & 221.69 \\
\textbf{Ours-RandomPose}
& \checkmark & \checkmark & \checkmark
& -- & -- & \textbf{9.28} & \underline{0.715} & \underline{0.809} & \underline{54.85} & \underline{235.02}
& \textbf{7.81} & \underline{0.584} & 0.875 & \textbf{39.87} & \textbf{155.92} \\
\textbf{Ours}
& \checkmark & \checkmark & \checkmark
& \textbf{0.098} & \textbf{0.34cm} & \underline{9.49} & \textbf{0.696} & \textbf{0.812} & \textbf{54.45} & \textbf{230.03}
& \underline{8.11} & \textbf{0.535} & \textbf{0.890} & \underline{41.09} & \underline{160.92} \\
\bottomrule
\end{tabular}
}
\end{table*}

\subsection{Implementation Details}
\label{sec:impl}
All models are trained on a single NVIDIA A100 GPU with a batch size of 16, using AdamW~\cite{adam} with learning rate $10^{-4}$, cosine decay, and 1k warmup steps; partially unfrozen encoder blocks use $10^{-5}$. The texture autoencoder trains for 60k steps and the joint stage for 100k steps, completing within 40 hours. Input images are resized to $256\times256$, texture latents are $64\times64\times8$, the fur map is $512\times512$ with 63 root-relative strand offsets per UV cell, and the decoded texture is $1024\times1024$. Inference completes in under 3 seconds using 50 Euler steps with CFG~\cite{ho2022classifier} scale 3.0, and outputs import directly into Blender.

\subsection{Applications}
\label{sec:applications}

AnimalLift reconstructs a canonical mesh, UV texture, and UV-aligned fur field from a single image, enabling animation-ready 3D animals. Because all meshes share topology and UVs, they can be transferred to a common rig for animation and pose transfer, including SMAL poses from images, videos, or animation clips. Fig.~\ref{fig:animation} shows auto-rigging and animation examples.

The UV-aligned fur field attaches strands to the deforming surface, supporting Blender grooming and physics effects such as collision, gravity, and wind. The shared UV space also enables aligned edits to color, fur length, and grooming style. Fig.~\ref{fig:application} shows simulation and texture/fur editing examples. Fig.~\ref{fig:fur_transfer} shows that fur geometry and texture maps can transfer across different animal classes, even when their appearances differ substantially.

\subsection{Evaluation Metrics}
\label{sec:metrics}

We evaluate reconstruction quality on two test sets: 640 real photographs for real-world generalization (Section~\ref{sec:main_cmp}) and a held-out synthetic split with ground-truth meshes and fur annotations. The metrics cover three aspects: geometry accuracy, rendered appearance, and explicit fur quality.

Geometry is measured on the synthetic test set using Chamfer-L1 distance (CD-L1)~\cite{fan2017point}. Since different methods use different scales and coordinate systems, all predicted meshes are aligned to the ground truth with similarity ICP~\cite{besl1992method} and normalized to a unit bounding box. For our canonical output, we directly compare the predicted canonical mesh with the ground-truth canonical mesh when computing CD-L1. 

For appearance, we uniformly render each reconstructed asset from six views and evaluate the renderings using NIQE~\cite{niqe}, LPIPS~\cite{lpips}, MV-CLIP~\cite{clip}, $FID_{\mathrm{CLIP}}$~\cite{heusel2017gans}, and Patch-FID$_{\mathrm{Inc}}$. NIQE measures reference-free naturalness, LPIPS measures perceptual similarity to the input photograph, and MV-CLIP measures semantic consistency between the input image and the multi-view renderings using CLIP features. $FID_{\mathrm{CLIP}}$ measures distribution-level realism against real animal photographs in CLIP feature space, while Patch-FID$_{\mathrm{Inc}}$ computes FID on foreground animal patches using Inception features to better capture local fur realism.
\subsection{Reconstruction Quality Comparison}
\label{sec:main_cmp}

For real-world generalization, we evaluate on 640 photographs from Oxford-IIIT Pet~\cite{oxford_pet} and Animals with Attributes 2~\cite{awa2}, covering dogs, cats, foxes, wolves, bears, and lions. All photographs are external to training; only ${\sim}20\%$ share a breed label with training and the remaining ${\sim}80\%$ are held-out combinations. For geometry and fur, we also evaluate on the held-out synthetic test split with ground-truth meshes and fur annotations.

We compare with representative image-to-3D and animatable animal reconstruction methods. Hunyuan-3D~2.1~\cite{hunyuan3d2025hunyuan3d} and Trellis 2~\cite{trellis} are general image-to-3D methods, Fauna-3D~\cite{li2024learning} is animal-specific, and BITE~\cite{ruegg2023bite} and AniMer~\cite{lyu2025animer} reconstruct SMAL-based animatable geometry without texture or fur. Since these methods produce different representations, Table~\ref{tab:main_cmp} compares both reconstruction quality and supported asset capabilities.

Quantitative results are reported in Table~\ref{tab:main_cmp}, with visual comparisons in Fig.~\ref{fig:comparsion2}. Our method reconstructs canonical assets, whereas several baselines reconstruct posed geometry aligned to the input image and do not support re-posing. We therefore report \textbf{Ours-RandomPose}, where our canonical output is re-posed using a randomly sampled SMAL pose from the dataset pose pool, to evaluate animation quality under articulated rendering. For BITE and AniMer, which do not predict texture, we render their meshes using our reconstructed texture.

Overall, AnimalLift achieves competitive reconstruction quality while uniquely producing structured canonical assets with explicit fur. Although image-space metrics such as LPIPS and MV-CLIP can favor methods that directly reconstruct the input pose, our method still achieves strong LPIPS and MV-CLIP scores, indicating good visual and semantic agreement with the input image. The lower Patch-FID further shows improved local fur realism compared with methods that represent fur only as surface texture, which often leads to oversmoothed or inconsistent high-frequency appearance. On the synthetic benchmark, AnimalLift obtains the best canonical geometry accuracy and reconstructs explicit fur with an L1-Hair error of 0.34\,cm at near real-world scale. Since prior methods do not output fur, they cannot be evaluated with this metric. The close performance of \textbf{Ours} and \textbf{Ours-RandomPose} also shows that animation does not significantly degrade visual quality. Because a single image leaves large portions of the surface unobserved, AnimalLift completes these occluded regions in the canonical UV space using category-level priors learned under fully-observed supervision, which is what allows the reconstructed canonical asset to be fully textured and groomed despite single-view input.

Beyond image-space scores, our canonical representation enables animation, grooming, editing, and simulation-ready assets, as demonstrated in the supplementary video. Beyond image-space scores, our canonical representation directly yields simulation-ready assets—a DCC-compatible base of shared-topology mesh, UV texture, and Blender curve-based fur with UV attributes and mesh bindings—that can be rigged, re-posed, groomed, and simulated under collision, gravity, and wind, then rendered without further conversion, as demonstrated in the supplementary video. Only hero-quality shots typically require additional artist cleanup.

\subsection{Ablation Study}
\label{sec:ablation}

We ablate our key design choices on the same 640 real-world photographs used in the main comparison. Each variant removes or replaces a single component while keeping all other settings fixed. We further include a variant that additionally applies flow matching to the fur map, to justify our choice of deterministic fur regression.

\begin{table}[!htbp]
\centering
\caption{Ablation study on the real-image benchmark. Each row removes or replaces one component of the full model. Best results are \textbf{bold}.}
\label{tab:ablation}
\resizebox{\linewidth}{!}{
\begin{tabular}{lccccc}
\toprule
Variant
& NIQE $\downarrow$
& LPIPS $\downarrow$
& MV-CLIP $\uparrow$
& FID$_{\text{CLIP}}$ $\downarrow$
& Patch-FID$_{\text{Inc}}$ $\downarrow$ \\
\midrule
w/o fur prediction   & 8.93 & 0.604 & 0.863 & 43.45 & 190.93 \\
w/o flow matching    & 10.02 & 0.648 & 0.774 & 49.99 & 210.32 \\
w/o CFG              & 9.21 & 0.603 & 0.832 & 46.52 & 187.32 \\
w/o color embed.     & 8.73 & 0.548 & 0.834 & 44.23 & 180.46 \\
w/o geometric reg.   & 8.32 & 0.551 & 0.874 & 43.11 & 173.49 \\
w/ flow matching (fur) & 8.59 & 0.594 & 0.884 & 42.69 & 162.14 \\ 
\midrule
Full model (ours)    & \textbf{8.11} & \textbf{0.535} & \textbf{0.890} & \textbf{41.09} & \textbf{160.92} \\
\bottomrule
\end{tabular}
}
\end{table}

As shown in Table~\ref{tab:ablation}, removing explicit fur prediction reduces local realism, increasing Patch-FID$_{\text{Inc}}$ from 160.92 to 190.93, confirming the importance of explicit strand geometry for realistic fur appearance. Replacing flow matching with direct regression causes the largest overall degradation, producing blurrier textures and increasing Patch-FID$_{\text{Inc}}$ to 210.32. Disabling classifier-free guidance (CFG) further reduces texture sharpness and semantic consistency, reflected by the drop in MV-CLIP and increase in Patch-FID$_{\text{Inc}}$. Removing the color embedding mainly decreases similarity between reconstructed appearance and the input animal, while removing geometric regularization causes moderate degradation across realism and perceptual metrics. Finally, applying flow matching to the fur map yields essentially no quality change while requiring substantially longer training, supporting deterministic fur regression.
\subsection{Robustness Test}
\label{sec:corruption}

\paragraph{Robustness to Image Corruption.} We evaluate robustness to degraded input images by applying four severe corruptions to the input image: brightness shift, JPEG compression, Gaussian noise, and motion blur. Specifically, brightness shift multiplies RGB intensity by $2.0$; JPEG compression uses quality $10$; Gaussian noise adds independent noise with $\sigma=40$ in the $[0,255]$ RGB space; and motion blur applies a horizontal box filter with kernel size $17$. 

\begin{table}[!htbp]
\centering
\caption{Robustness evaluation under severe input corruptions. Lower is better for all metrics except MV-CLIP (higher is better).}
\label{tab:corruption}
\resizebox{\linewidth}{!}{
\begin{tabular}{lccccc}
\toprule
Input 
& NIQE$\downarrow$ 
& LPIPS$\downarrow$ 
& MV-CLIP$\uparrow$
& FID$_{\mathrm{CLIP}}\downarrow$ 
& Patch-FID$_{\mathrm{Inc}}\downarrow$ \\
\midrule
Clean & 8.110 & 0.5350 & 0.8908 & 41.09 & 160.92 \\
Brightness & 9.035 & 0.5684 & 0.8605 & 50.15 & 163.21 \\
JPEG Comp. & 9.251 & 0.5621 & 0.8756 & 55.66 & 161.20 \\
Gaussian Noise & 8.274 & 0.5606 & 0.8857 & 52.24 & 160.74 \\
Motion Blur & 8.530 & 0.5596 & 0.8904 & 40.43 & 159.85 \\
\midrule
Avg. Corrupted & 8.772 & 0.5627 & 0.878 & 49.62 & 161.25 \\
$\Delta$ vs. Clean & +0.662 & +0.0277 & -0.0128 & +8.53 & +0.33 \\
\bottomrule
\end{tabular}
}
\end{table}

\begin{figure}[!htbp]
  \includegraphics[width=1\linewidth]{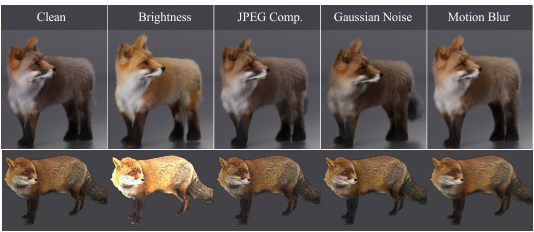}
  \caption{Robustness visualization under severe input corruptions. Each column shows a corrupted \emph{input} image (inset) and our reconstruction from it; no competing method is shown here.}
  \Description{Two rows compare reconstructions from clean, altered-brightness, JPEG-compressed, Gaussian-noisy, and motion-blurred input images. Each rendering includes its input image as an inset.}
    \label{fig:image_quality}
\end{figure}

Quantitative results are reported in Table~\ref{tab:corruption}, with visual examples shown in Fig.~\ref{fig:image_quality}. Overall, all corruptions cause only minor degradation, indicating that the model relies on robust geometric and appearance priors rather than low-level image details. Brightness shift produces the largest performance drop because it significantly alters the global color distribution. However, the reconstructed textures still maintain plausible fur colors and clear grooming details instead of becoming overexposed like the corrupted inputs.

\paragraph{Viewpoint and Pose Sensitivity}

We further evaluate robustness under varying viewpoints and articulated poses. For viewpoint evaluation, we generate 90 images using Qwen-Image~\cite{wu2025qwen}, covering 30 animal instances rendered from frontal ($0^\circ$), oblique ($45^\circ$), and side ($90^\circ$) views. For pose evaluation, we generate 90 image pairs showing the same animal under different articulations. Pose differences are estimated using BITE and grouped into low, medium, and high deformation levels according to joint displacement magnitude.

\begin{figure}[!htbp]
  \includegraphics[width=1\linewidth]{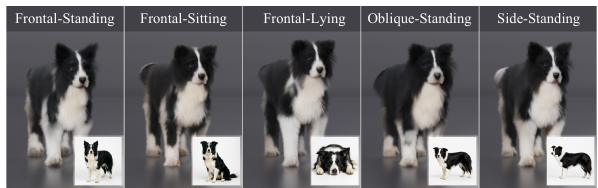}
  \caption{
  Robustness visualization across different viewpoints and poses.
  The rendered images are quite consistent, despite input image variations (inset).
  }
  \Description{Five columns show reconstructions from frontal standing, frontal sitting, frontal lying, oblique standing, and side standing animal images, with the input views shown as insets.}
    \label{fig:different_view}
\end{figure}

Table~\ref{tab:view_pose} and Fig.~\ref{fig:different_view} show that the model remains stable across different viewpoints and pose changes with only minor metric variation. Even under side views and large articulation changes, the reconstructed geometry, texture, and fur appearance remain coherent without substantial degradation.

\begin{table}[!htbp]
\centering
\caption{Robustness evaluation under viewpoint and pose changes. Lower is better for all metrics except MV-CLIP (higher is better).}
\label{tab:view_pose}
\resizebox{\linewidth}{!}{
\begin{tabular}{lccccc}
\toprule
Input 
& NIQE$\downarrow$ 
& LPIPS$\downarrow$ 
& MV-CLIP$\uparrow$
& FID$_{\mathrm{CLIP}}\downarrow$ 
& Patch-FID$_{\mathrm{Inc}}\downarrow$ \\
\midrule
Frontal-L1-Deform &8.220& 0.647&0.861&51.331&176.258\\
Oblique &8.433&0.650&0.881&50.539&176.613 \\
Side & 8.564&0.650&0.850&52.840&173.059 \\
L2-Deform &8.434&0.654&0.863&52.371&173.605\\
L3-Deform &8.174&0.639&0.862&53.402&176.923\\
\bottomrule
\end{tabular}
}
\end{table}

\input{sec/final_comparisons}

\subsection{Generalization to Held-out Breeds and Identities}
Because the 640 real photographs used above contain breed and appearance combinations that are absent from our prototype templates, they also serve as an out-of-distribution generalization test. Rather than retrieving a fixed template, AnimalLift recombines learned shape and appearance priors, so for unseen breeds it preserves distinguishing traits such as ear shape, muzzle length, body proportion, and local fur density (Fig.~\ref{fig:demo2}). We further examine identity preservation within a single species. For clearly distinct individuals, for example two wolves with different coat patterns and facial markings, the model recovers separate texture and facial features; for near-identical individuals, the finest identity-specific cues such as broken strands or scars are less faithfully reproduced. This behavior is consistent with our intentional canonical-appearance design (Section~\ref{sec:conclusion}): the network reconstructs a clean, riggable groom and defers transient, instance-specific detail to downstream editing. Importantly, across these cases the reconstructed geometry and shared topology remain animation-compatible, indicating that generalization to new breeds and identities does not compromise the structural consistency required for downstream rigging and simulation.

\subsection{User Study}
We conduct a blind expert study to evaluate the production-readiness of reconstructed 3D animal assets. Ten experts evaluate five randomly selected animal cases using Blender files. This study targets DCC usability with experienced artists under standardized workflows rather than crowd-sourced perceptual preference, which makes expert feedback lower-variance and more task-relevant. For each case, participants compare three anonymized methods in randomized order and assess visual quality, mesh/asset usability, static-render cleanup effort, walking-animation preparation effort, and the highest quality level achievable after one hour of cleanup. The latter is evaluated using a four-point production scale:
\begin{enumerate}[label=\arabic*)]
\item not usable,
\item distant-shot use only,
\item mid-range use in games or animation, and
\item close- or mid-range use in animation or other digital production workflows.
\end{enumerate}

Our method is consistently preferred for both visual quality and mesh/asset usability, receiving 50/50 votes for each criterion. It also requires substantially less cleanup effort: 1.7 hours on average for static rendering, compared to 3.8 hours for Hunyuan-3D~2.1 and 6.8 hours for Trellis 2. For walking animation, our method requires 1.5 hours on average, while Hunyuan-3D~2.1 requires approximately 18.2 hours, and Trellis 2 is generally rated as requiring more than two days or being impractical for production use. After one hour of cleanup, our method achieves an average quality level of 4.0/4.0, compared to 2.58 for Hunyuan-3D~2.1 and 1.72 for Trellis 2. Detailed rating distributions are shown in Fig.~\ref{fig:user_study}, with full study details provided in the supplementary material Section 3.

%% file: sec/final_comparisons.tex
\begin{figure}[!t]
    \centering
    \includegraphics[width=1\linewidth]{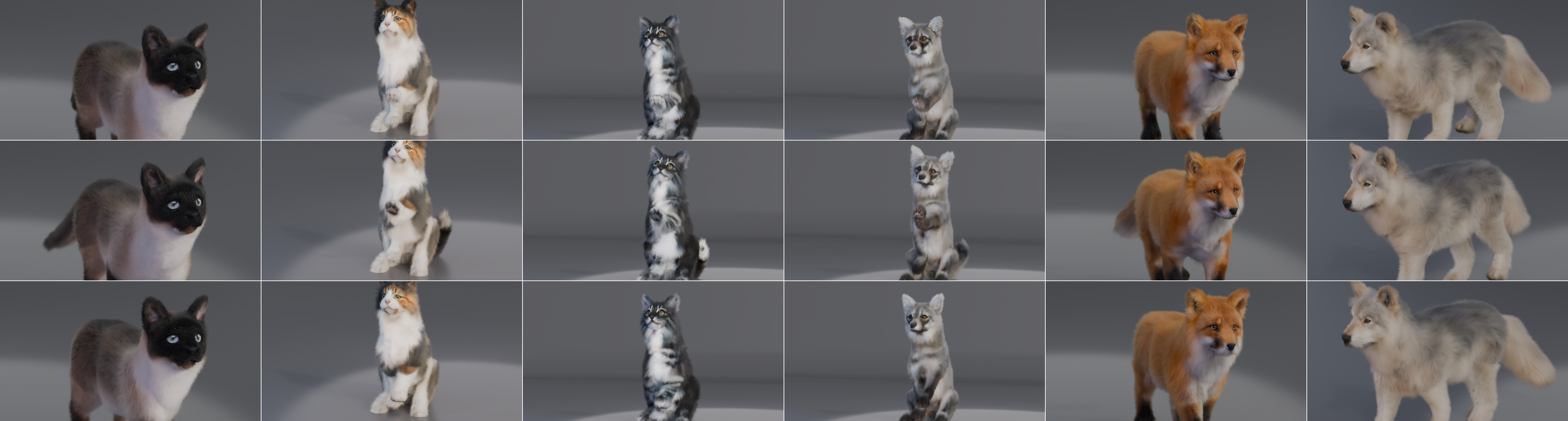}
    \caption{\textbf{Automatic rig transfer for animal animation.}
Our program transfers rigging and template actions across furry animal models, producing natural motion and fur flow. }
    \Description{A grid shows six reconstructed furry animals in three successive animated poses, illustrating transferred rigging and motion.}
    \label{fig:animation}
\end{figure}

%% file: sec/6_conclusion.tex
\section{Conclusion and Discussion}
\label{sec:conclusion}


\paragraph{Conclusion.}
We present AnimalLift, a framework that introduces a structured canonical representation for reconstructing animatable 3D animal assets with explicit fur from a single image. Our results bridge learning-based reconstruction and graphics pipelines for content creation and animation. By modeling geometry, texture, and fur within a shared UV-aligned representation, our framework preserves dense correspondence and supports downstream editing, animation, grooming, and simulation-compatible rendering workflows. We hope this work provides a foundation for future research on structured animal modeling and fur simulation.


\paragraph{Limitations and Future Work.}
Our method assumes a shared topology and is currently limited to a family of quadruped animals. Our method assumes a shared topology and is currently limited to a family of quadruped animals. Shared topology constrains mesh connectivity and correspondence rather than geometry, so inter-species differences in muzzle, ear, and limb proportions remain captured by the predicted per-vertex offsets. Reconstructions can nonetheless appear more uniform than the photographed individual, partly by design—we recover a clean canonical groom rather than transient states such as wetness, dirt, or clumping, favoring editability over photo-specific individuality—and partly as a limitation, since fine identity cues such as scars or broken strands are not always recovered with the diversity of our current templates.
In addition, reconstructing from a single image introduces inherent ambiguity, particularly for occluded regions: the category priors used to complete them (Section~\ref{sec:main_cmp}) yield plausible rather than ground-truth predictions for unobserved surfaces. While this setting enables scalable asset creation from widely available images, applications requiring high-fidelity reconstruction of specific instances may benefit from multi-view input. The artist-authored templates are a one-time cost; scalability could be further improved by adapting canonical assets from image-to-3D models or weak supervision on in-the-wild images. Extending our framework to incorporate multi-view or video observations is a promising direction.